\documentclass[nonacm]{acmart}
\pdfoutput=1

\AtBeginDocument{%
  }
\usepackage{graphicx} 

\usepackage{url}            
\usepackage[utf8]{inputenc} 
\usepackage[T1]{fontenc}    
\usepackage{xcolor}  
\usepackage{hyperref}       
\hypersetup{colorlinks=true,linkcolor=blue,allcolors=blue,linktocpage=true}

\usepackage{booktabs}       
\usepackage{amsfonts}       
\usepackage{nicefrac}       
\usepackage{microtype}      
\usepackage{amsmath}
\usepackage{mathtools}
\usepackage{amsthm, bbm, bm}
\usepackage{thm-restate, thmtools, enumitem}

\usepackage{algorithm}
\usepackage{soul, enumitem}
\usepackage{multicol}
 \usepackage{algpseudocode} 
\newtheorem{lemma}{Lemma}

\newtheorem{theorem}{Theorem}
\newtheorem{assumption}{Assumption}
\newtheorem{definition}{Definition}

\newtheorem{remark}{Remark}

\usepackage[normalem]{ulem}
\usepackage{balance} 

\usepackage[capitalize,noabbrev]{cleveref}

\usepackage{subfigure}

\DeclarePairedDelimiterX\brk[2]{\langle}{\rangle}{#1\,,\,#2} 
\DeclarePairedDelimiterX\Set[2]{\{}{\}}{#1 \;;\; #2} 

\newcommand{\argmax}{\operatornamewithlimits{arg\,max}}

\newcommand\independent{\protect\mathpalette{\protect\independenT}{\perp}}
\def\independenT#1#2{\mathrel{\rlap{$#1#2$}\mkern2mu{#1#2}}}
\newcommand\indep\independent

\newcommand{\EE}{\mathbb{E}}

\newcommand{\NN}{\mathbb{N}}

\newcommand{\PP}{\mathbb{P}}

\newcommand{\RR}{\mathbb{R}}

\newcommand{\Acal}{\mathcal{A}}

\newcommand{\Ccal}{\mathcal{C}}

\newcommand{\Ecal}{\mathcal{E}}

\newcommand{\Scal}{\mathcal{S}}

\newcommand{\Xcal}{\mathcal{X}}

\newcommand{\Bal}{\begin{align}}
\newcommand{\Eal}{\end{align}}
\newcommand{\Beq}{\begin{equation}}
\newcommand{\Eeq}{\end{equation}}
\newcommand{\Bit}{\begin{itemize}}
\newcommand{\Eit}{\end{itemize}}
\newcommand{\Ben}{\begin{enumerate}}
\newcommand{\Een}{\end{enumerate}}
\newcommand{\Ba}{\begin{array}}
\newcommand{\Ea}{\end{array}}

\newcommand{\Bvec}{\left(\begin{array}{c}}
\newcommand{\Evec}{\end{array}\right)}

\newcommand{\Bmat}{\left(\begin{array}}
\newcommand{\Emat}{\end{array}\right)}

\newcommand{\Bol}{\begin{outline}}
\newcommand{\Eol}{\end{outline}}

\begin{document}

\title[SNPL: Simultaneous Policy Learning and Evaluation for Safe Multi-Objective Policy Improvement]{SNPL: Simultaneous Policy Learning and Evaluation for Safe Multi-Objective Policy Improvement}
\author{Brian M Cho}
\authornote{Work done during an internship at Meta.}
\authornotemark[0]
\affiliation{
    \institution{Cornell University}
    \city{New York}
    \state{New York}
    \country{USA}
}

\author{Ana-Roxana Pop}
\affiliation{
    \institution{Meta}
    \city{New York}
    \state{New York}
    \country{USA}
}

\author{Ariel Evnine}
\affiliation{
    \institution{Meta}
    \city{San Francisco}
    \state{California}
    \country{USA}
}

\author{Nathan Kallus}
\affiliation{
    \institution{Cornell University, Netflix}
    \city{New York}
    \state{New York}
    \country{USA}
}

\renewcommand{\shortauthors}{Brian M Cho et al.}


\begin{abstract}

To design effective digital interventions, experimenters face the challenge of learning decision policies that balance multiple objectives using offline data. Often, they aim to develop policies that maximize \emph{goal} outcomes, while ensuring there are no undesirable changes in \emph{guardrail} outcomes. To provide credible recommendations, experimenters must not only identify policies that satisfy the desired changes in goal and guardrail outcomes, but also offer probabilistic guarantees about the changes these policies induce. In practice, however, policy classes are often large, and digital experiments tend to produce datasets with small effect sizes relative to noise. In this setting, standard approaches such as data splitting or multiple testing often result in unstable policy selection and/or insufficient statistical power. In this paper, we provide safe noisy policy learning (SNPL), a novel approach that leverages the concept of algorithmic stability to address these challenges. Our method enables policy learning while simultaneously providing high-confidence guarantees using the \emph{entire dataset}, avoiding the need for data-splitting. We present finite-sample and asymptotic versions of our algorithm that ensure the recommended policy satisfies high-probability guarantees for avoiding guardrail regressions and/or achieving goal outcome improvements. We test both variants of our approach approach empirically on a real-world application of personalizing SMS delivery. Our results on real-world data suggest that our approach offers dramatic improvements in settings with large policy classes and low signal-to-noise across both finite-sample and asymptotic safety guarantees, offering up to 300\% improvements in detection rates and 150\% improvements in policy gains at significantly smaller sample sizes.
\end{abstract}


\keywords{Differential Privacy, Safe Policy Learning, Statistical Efficiency, Multiple Testing, Selective Inference}



\maketitle



\section{Introduction}

Many decision-making scenarios involve the optimization of multiple, potentially conflicting objectives. Experimenters strive to create policies that achieve the best possible results for their primary objectives (goal outcomes) while ensuring that they do not negatively impact other important objectives (guardrail outcomes). For instance, technology companies may seek to increase user retention by targeting SMS notifications to highly elastic users, while avoiding large increases in overall SMS delivery costs. In such scenarios, the objective is to learn policies that maximize goal outcomes (e.g., retention), while simultaneously ensuring that guardrail outcomes (e.g., SMS costs) remain within acceptable limits. 

While existing works for learning multi-objective policies focus on achieving pareto-optimal solutions for multiple objectives \cite{mutli_obj}, learning weighted outcome functions to maximize a long-term outcome \cite{jeunen2024multiobjectiverecommendationmultivariatepolicy}, or balancing objectives while learning relative preferences \cite{yang2019generalizedalgorithmmultiobjectivereinforcement}, they do not provide statistical guarantees on the learned policy's quality. In industries such as technology \cite{taddy2015nonparametricbayesiananalysisheterogeneous} or healthcare \cite{liao2019personalizedheartstepsreinforcementlearning}, the collected data exhibits \emph{low} signal-to-noise ratios (SNR), where average changes in outcomes are relatively small compared to the variance of their estimates. In addition, policymakers and experimenters often must select among a large number of candidate policies. In such settings, policymakers may choose to adopt learned policies only if they can be deemed as \emph{safe} \cite{cho2024cspimtcalibratedsafepolicy, scholl2022safepolicyimprovementapproaches, laroche2019safepolicyimprovementbaseline} - i.e. satisfying all desired outcome changes with a high probability guarantee. This notion of safety ensures proposed changes to existing policies are not byproducts of noise and reflect the policy changes post-deployment with high probability.

Standard approaches for providing these guarantees in a statistically valid manner include data-splitting, where learning and statistical testing are conducted on disjoint subsets of the data, and multiple testing corrections, which conduct hypothesis tests on the entire set of considered policies. These solutions are imperiled in the low SNR regime with a large number of candidate policies. Data-splitting can result in different results across the two partitions: as a result, the selected policy, which may be estimated to satisfy the desired high-probability guarantees on the learning split, often fails to provide these guarantees when evaluated on the test split. On large candidate policy sets, multiple testing corrections suffer from weak statistical power, leading to conservative guarantees that are poorly suited for downstream decision-making.

In this work, we overcome the limitations of existing approaches in the low SNR regime with a large, discrete candidate policy class. We focus on offline policy learning from data collected previously from an experimental study. Our approach, called \emph{safe noisy policy learning} (SNPL), ensures the returned policy satisfies probabilistic guarantees for \emph{all} changes to guardrail and/or goal policy values relative to a pre-specified baseline policy. We provide two variants of our approach, which provide either finite-sample or asymptotic probabilistic guarantees under nonparametric assumptions. Both variants leverage the framework of algorithmic stability \cite{dwork_2006} to provide \emph{simultaneous} policy learning and inference, avoiding the inefficiencies of data-splitting \cite{pmlr-v37-thomas15} or multiple testing corrections over the whole candidate set \cite{Dunn_1961}. Our real-world experiments show that our approach (i) improves the detection rate of policies satisfying our guardrail constraints, (ii) maintains the desired error level for safety, and (iii) leads to larger expected gains in the goal outcome relative to other approaches with the same guarantees.

\paragraph{Outline} In Section \ref{sec:related_work}, we provide an overview of related works for offline policy learning/evaluation, selective inference, algorithmic stability, and safe policy learning. 
Section \ref{sec:setup} outlines our assumptions and problem formulation. We define the notion of safety, introduce the concept of algorithmic stability, and show how to construct joint confidence intervals for policy values. In Section \ref{sec:p1}, we introduce our safe policy learning approach called \textit{safe noisy policy learning}, or SNPL, and provide theoretical results demonstrating its safety. Lastly, in Section \ref{sec:experiments}, we empirically validate our approaches on a real-world application for SMS delivery personalization from a large consumer technology company. Our results demonstrate drastically improved detection rates and average gains for the goal outcome, showing the value of our approach in real-world settings.


\section{Related Work}\label{sec:related_work}
Our work draws upon four distinct areas of study: (i) offline policy learning and evaluation, (ii) post-selection inference, (iii) algorithmic stability (or differential privacy), and (iv) safe policy learning approaches. We provide an overview of these topics, highlighting the connections between our work and each of these fields. 

\paragraph{Offline Policy Learning and Evaluation} Offline policy learning \cite{athey_wager_policy_learning, nguyentang2022offlineneuralcontextualbandits, pmlr-v206-yang23f} and evaluation \cite{uehara2022reviewoffpolicyevaluationreinforcement, guo2024distributionallyrobustpolicyevaluation, NEURIPS2021_eff30581, zhan2021offpolicyevaluationadaptiveweighting} are fundamental topics in the bandit and reinforcement learning literature. The goal of offline policy learning is to learn a policy that maximizes an objective function, often specified as the average of a single outcome, using a historical (or given) dataset. In contrast, the goal of offline policy evaluation is to conduct inference on the value (e.g. average of outcomes) of a fixed policy, using a given dataset. While each problem has been studied extensively separately, few works consider the goal of \emph{simultaneously} learning a policy and evaluating this learned policy on the same dataset. The most ubiquitous approach to solving the selection problem is to partition the data into two distinct sets \cite{cox1975note, rubin2006method, ignatiadis2016data, cho2024cspimtcalibratedsafepolicy}, where one set is used to learn the policy and the other to evaluate the quality of this policy.
The closest approach in aim to our contributions is the CRAM method \cite{jia2024crammethodefficientsimultaneous}, which enables simultaneous learning and evaluation over most of the data. However, this approach is designed specifically for a single outcome, rather than multiple outcomes (e.g., goal and guardrails). Furthermore, CRAM only enables the construction of confidence intervals, without ensuring that the final policy satisfies any high-probability guarantees for desired changes to multiple outcomes.

\paragraph{Post-Selection Inference} To avoid data-splitting, a general approach for obtaining valid inference (such as confidence intervals or probabilistic guarantees) on data-driven policy selection involves applying multiple testing corrections \cite{Dunn_1961} across the entire set of candidate policies. By correcting for \emph{all} possible policy choices, this approach suffers from weak guarantees, as it unnecessarily provides inference for policies ultimately not selected by the learning algorithm. To address these drawbacks, the selective inference literature \cite{selective_inf, Lee_2016} provides methods for providing valid inference for a data-driven policy choice under parametric assumptions. In nonparametric settings, recent works \cite{jia2024crammethodefficientsimultaneous} use martingale difference arguments to sequentially learn policies based on multiple splits, retraining the policy with the number of splits. Our approaches avoid both (i) parametric assumptions and (ii) computationally intensive procedures (e.g. sequential retraining) by leveraging \emph{stable} \cite{bassily2015algorithmicstabilityadaptivedata} algorithms for policy selection, which allow for simultaneous policy learning and evaluation (i.e. probabilistic guarantees on a specific learned policy) with a single pass of the whole data.

\paragraph{Algorithmic Stability for Post-Selection Inference} 
The framework of algorithmic stability \cite{bassily2015algorithmicstabilityadaptivedata}, or differential privacy literature \cite{dwork_2006, dwork_textbook}, allows for corrections to inference procedures intended for a \emph{fixed} policy by limiting the sensitivity of a learning algorithm to its input data. While these corrections have been applied to problems such as model selection \cite{zrnic2022postselectioninferencealgorithmicstability} and causal discovery \cite{gradu2024validinferencecausaldiscovery}, our work is the first to leverage algorithmic stability for simultaneous policy learning and evaluation. Our work not only provides valid inference post policy selection, but further leverages the notion of stability for safe policy learning in both finite-sample and asymptotic regimes.

\paragraph{Safe Policy Learning}\label{disc:baselines} Our work differs from the existing literature for safe policy improvement in multiple ways. Existing work on safe policy learning \cite{scholl2022safepolicyimprovementapproaches, pmlr-v37-thomas15, cho2024cspimtcalibratedsafepolicy} only considers a single outcome, with the goal of returning a policy that has a higher average outcome relative to a baseline with high probability. In the reinforcement learning (RL) setting, existing approaches  \cite{scholl2022safepolicyimprovementapproaches, laroche2019safepolicyimprovementbaseline} use computationally expensive procedures to conduct safety tests based on boostrapping. However, this approach is strictly limited to discrete state (or context) variables and focuses on the general RL setting. In our work, we do not make any assumptions on our context variables, and focus specifically on a simpler contextual bandit treatment policy\footnote{In particular, we assume that the context distribution does not depend on the actions taken prior, reflecting common assumptions in the literature \cite{athey_wager_policy_learning, kitagawa_policy_learning}.}. Beyond bootstrapping methods, other existing approaches leverage data-splitting \cite{pmlr-v37-thomas15, cho2024cspimtcalibratedsafepolicy} in order to conduct safe policy learning. As mentioned before, our approach differs by using the entire dataset for both learning and inference, and provide guarantees for multiple outcomes rather than just the goal outcome.



\section{Problem Setup and Preliminaries} \label{sec:setup}

We observe data consisting of $n$ experimental observations $\{O_i\}_{i=1}^n$, where each observation consists of covariates, treatment, and outcomes, $O_i = (X_i, A_i, Y_i) \in \Xcal \times [K] \times [0,1]^{d_Y}$. The space $\Xcal$ is a general measurable space (i.e. can be discrete, continuous, or both). We consider a discrete set of treatments $[K] = \{1,...K\}$, and a bounded vector of outcomes $Y$, where $Y_{j}$ denotes the $j$-th outcome of interest. We assume that $\{O_i\}_{i=1}^n$ are independently and identically distributed, with the common distribution $P = P_X \times P_{A|X} \times P_{Y|A,X}$. We denote the propensity score function as $e(k,x) = P(A=k | X=x)$ and the conditional expectation for the $j$-th outcome as $\mu_j(k,x) = \EE_{P}[Y_j|A=k, X=x]$.  We make no assumptions on the form of distributions $P_X$ and $P_{Y|A,X}$, corresponding to the nonparametric setting. Given that the data is collected from an experimental setting, we assume propensities $e(k,x)$ are known. Additionally, on the propensity score function $e(k,x)$, we assume strict positivity, a common assumption in both causal inference \cite{dml} and the off-policy evaluation \cite{uehara2022reviewoffpolicyevaluationreinforcement} that restricts $P_{A|X}$ in the following manner.

\begin{assumption}[Strict Positivity]\label{assump:strict_positivty}
    We assume for all treatments $k \in [K]$, all covariates $x \in \Xcal$, 
    $e(k,x)\geq c$ for some constant $c \in (0,1)$. 
\end{assumption}

We define a policy $\pi: \Xcal \rightarrow \Delta^K$ as a mapping from the covariate space $\Xcal$ to a distribution over $[K]$. We denote $\pi(x,a)$ as the probability of selecting action $A=a$ given covariate $X=x$ under policy $\pi$, with $\sum_{a \in [K]}\pi(x,a) = 1$ for all possible values of $x$. In this work, we consider a finite set of policies $\Pi$ with cardinality $|\Pi|$, which includes a baseline policy $\pi_0$. The baseline policy $\pi_0$ reflects some status-quo policy. We note that $\pi_0$ does not necessarily need to be the policy used to obtain data $\{O_i\}_{i=1}^n$. We define $V_j(\pi)$, the $j$th policy value of a given policy $\pi \in \Pi$, as the average value of outcome $Y_j$ under the distribution induced by the policy $\pi$. 

\begin{definition}[Policy Value for Outcome $j$]
    For a given policy $\pi \in \Pi$, we define the policy value of $\pi$ for outcome $j$ as the average value of outcome $Y_j$ under policy $\pi$, i.e. 
    \begin{equation}\label{eq:policy_value}
        V_j(\pi) = \EE_{P_X}\left[ \sum_{k \in [K]} \pi(k, X) \ \mu_j(k,X) \right].
    \end{equation}
\end{definition}


\subsection{Problem Statement}
Our work focuses on the goal of \emph{safe} policy learning, considering both finite sample sizes, where $n$ can take any value, as well as asymptotic settings, where the sample size $n \rightarrow \infty$. We first begin by defining our goal and guardrail outcomes. Let $g \in [d_Y]= \{1,...,d_Y\}$ denote the goal outcome index, and $\Scal \subseteq [d_Y]$ denote the indices for the guardrail outcomes. Let $w \in \RR^{|\Scal|}$ denote a vector of guardrail weights, where $w_j$ corresponds to the $j$-th entry of $w$ and $|\Scal|$ denotes the number of indices in $\Scal$. Our goal is to construct a policy learning algorithm that aims to maximize $V_g(\pi)$ while satisfying $(\Scal, \alpha, w)$-safety, defined in Definition \ref{defn:asymp_safe_alg}. 

\begin{definition}[$(\Scal,\alpha, w)$-Safe Algorithm]\label{defn:asymp_safe_alg}
Let an algorithm $\Acal: \left( \{O_i\}_{i=1}^n, \Scal, \alpha, w \right) \rightarrow \Pi$ be a mapping from the observed data, the set of sensitive outcomes $\Scal$, $\alpha$, and guardrail weights $w \in \{\RR_- \cup 0\}^{|\Scal|}$ to a policy $\pi \in \Pi$, denoted as $\hat\pi_{\Acal}$. We say that an algorithm $\Acal$ is $(\Scal, \alpha, w)$-safe if the returned policy, $\hat\pi_{\Acal}$, obtains at least $1-w_i$ proportion of the baseline policy value for all outcome $j \in \Scal$ with at least probability $1-\alpha$ for all $n \in \NN$, where $\NN$ is the set of positive integers, i.e.,  
\begin{equation}
    \forall n, \ P\left(
                  \exists j \in \Scal : V_j(\hat\pi_{\Acal}) - (1+w_j)V_j(\pi_0)  <  0\right) \leq \alpha.
\end{equation}
Additionally, we say that an algorithm $\Acal$ is $(\Scal, \alpha, w)$-safe asymptotically if it satisfies $(\Scal, \alpha, w)$-safety as $n \rightarrow \infty$, i.e, 
    \begin{equation}
                \limsup_{n \rightarrow \infty} P\left(
                  \exists j \in \Scal : V_j(\hat\pi_{\Acal}) - (1+w_j)V_j(\pi_0)  <  0 \right) \leq \alpha.
    \end{equation}
\end{definition}
An $(\Scal, \alpha, w)$-safe algorithm $\Acal$ only returns a policy $\hat\pi_{\Acal}$ if it can ensure that $\hat\pi_{\Acal}$'s policy values achieves at least $(1+w_j)$ proportion of the baseline policy value for \emph{all} guardrail outcomes $j \in \Scal$. The weights $w_j$ indicate the desired level of change in the guardrail outcomes. The nonpositive values of $w_j$ indicate that we tolerate $w_j$-proportional decreases for the $j$th policy value from the baseline policy $\pi_0$ with high probability. When $w_j=0$, this indicates that we tolerate no regression in the guardrail relative to the baseline policy value for outcome $j$. Our definition of safety is a strict generalization of the standard definition for safety in the policy learning literature \cite{pmlr-v37-thomas15, cho2024cspimtcalibratedsafepolicy}: one can set $\Scal = \{g\}$ and $w_g = 0$ to recover the standard definition in the single outcome setting. We note that $g$ and $\Scal$ need not be disjoint, so one may include $g$ in $\Scal$ with $w_g = 0$ in order to obtain high-probability guarantees that policy values for goal outcome do not decrease with our safe policy learning approach. 

\begin{remark}[Why asymptotic guarantees?] 

One may wonder why we introduce an asymptotic notion $(\Scal, \alpha, w)$-safety in Definition \ref{defn:asymp_safe_alg}, as any algorithm that satisfies $(\Scal, \alpha, w)$-safety necessarily satisfies $(\Scal, \alpha, w)$-asymptotic safety. We introduce the relaxed notion of safety in order to leverage more powerful safety testing methods with only asymptotic guarantees. In large sample regimes, many finite-sample inference tools (and likewise, safety guarantees) are too conservative in practice for constructing confidence sets and safety guarantees \cite{cho2024cspimtcalibratedsafepolicy, pmlr-v37-thomas15}. We show the benefits of focusing on the asymptotic definition, which provides improved power and policy gains, in Section \ref{sec:experiments} below.
\end{remark}

Our work provides two algorithms: an $(\Scal, \alpha, w)$-safe algorithm with conservative performance, and an $(\Scal, \alpha, w)$-asymptotically safe algorithm that performs better in practice. Both of our approaches build upon two key tools: (i) an $\epsilon$-stable selection algorithm for constructing the subset of policies $\widehat{\Pi} \subseteq \Pi$ and (ii) joint confidence lower bounds over guardrail outcomes $j \in \Scal$ for all policies $\pi \in \widehat{\Pi}$ selected in an $\epsilon$-stable manner.

In the following section, we first focus on the latter tool. We introduce joint lower confidence bounds intended for a pre-specified policy subset $\widetilde\Pi \subseteq \Pi$, define the notion of $\epsilon$-stability, and show how to construct joint lower confidence bounds for a policy set $\widehat{\Pi}$ obtained with an $\epsilon$-stable data-driven selection process.

\section{Joint Post-Selection Confidence Bounds} \label{sec:p1}
We provide approaches for constructing joint post-selection confidence bounds, given that selected policy set $\widehat\Pi$ satisfies the condition of $\epsilon$-stability in Definition \ref{defn:stable} below. We first begin by introducing approaches for constructing finite-sample and asymptotic joint confidence intervals that provide uniform coverage guarantees over all policy values $V_j(\pi)$ for all $\pi$ in a given set of policies $\widetilde{\Pi}$.  We begin with our finite-sample joint confidence intervals.

\subsection{Finite-Sample Joint Lower Bounds}

To construct finite-sample confidence intervals, we leverage the inverse-propensity weighted estimator, provided in Definition \ref{defn:ipw}. 

\begin{definition}[Inverse Propensity Weighted (IPW) Estimator]\label{defn:ipw}
    Let $\psi_{jk}^{IPW}(O; e) = \frac{\mathbf{1}[A=k]Y_j}{e(k, X)}$, and $\psi_j^{IPW}(O, \pi;e) = \sum_{k \in K} \pi(k, X) \psi_{jk}^{IPW}(O;e)$. Then, the IPW estimator for the $j$th policy value under policy $\pi$ is
    \begin{equation}
        \hat{V}^{IPW}_j(\pi) = \frac{1}{n} \sum_{i=1}^n \psi_j^{IPW}(O_i, \pi; e), 
    \end{equation}
    where $e$ denotes the propensity score function for data collection. 
\end{definition}

Using our IPW estimates for policy value $j$ for a fixed policy $\pi$, we obtain lower joint confidence bounds in Lemma \ref{lem:finite_sample_joint_ci}.

\begin{lemma}[Joint Finite Sample Lower Confidence Bounds]\label{lem:finite_sample_joint_ci}
Let $\widetilde{\Pi}$ denote a fixed set of policies, and let $\Scal$ be the set of guardrail outcomes. Define lower bound confidence sets for policies $\pi \in \widetilde{\Pi}$ as $\hat{\Ccal}^{IPW}(\pi,\alpha)$, where the $\hat{\Ccal}_j^{IPW}(\pi,\alpha)$, the $j$th lower bound of $\hat{\Ccal}^{IPW}(\pi,\alpha)$, is defined as follows:
\begin{align}
    \hat\Ccal_j^{IPW}(\pi, \alpha) &= \left(\hat{V_j}^{IPW}(\pi) - (1+w_j)\hat{V}_j^{IPW}(\pi_0)\right) \\
    & - \left(\hat{\sigma}_j(\pi)\sqrt{\frac{2\log\left(\frac{3|\widetilde{\Pi}||\Scal|}{2\delta}\right)}{n}} + \frac{3R_j\log\left(\frac{3|\widetilde{\Pi}||\Scal|}{2\delta}\right)}{n}  \right)
\end{align}
where $|\widetilde{\Pi}|$ is the cardinality of $\widetilde{\Pi}$, $R_j = (2+w_j)/c$, and $\hat\sigma_j^2(\pi)$ is the empirical variance defined as follows:
\begin{align*}
     \hat\sigma_j^2(\pi) = \frac{1}{n}\sum_{i=1}^n &\biggr{(} \left( \psi^{IPW}_j(O, \pi;e) - (1+w_j)\psi^{IPW}_j(O, \pi_0;e) \right) \\
    &- \left(\hat{V_j}^{IPW}(\pi) - (1+w_j)\hat{V}_j^{IPW}(\pi_0)\right)\biggr{)}^2.
\end{align*}

Then, for all $n \in \NN$, all $\pi \in \Pi$, $\hat{C}_j^{IPW}(\pi, \alpha)$ are joint $1-\alpha$ lower confidence bounds for $V_j(\pi) - \left(1+w_j\right)V_j(\pi_0)$, i.e., for all $n \in \NN$,
\begin{equation}
    P\left(\exists \pi \in \widetilde{\Pi}, \ j \in \Scal : \left(V_j(\pi) - \left(1+w_j\right)V_j(\pi_0) \right)< \hat{\Ccal}_j^{IPW}(\pi, \alpha) \right) \leq \alpha.
 \end{equation}
    
\end{lemma}
These finite-sample bounds pair an empirical Bernstein inequality \cite{Mnih2008EmpiricalBS} with a union bound in order to obtain lower confidence bounds. While the lower bounds in Lemma \ref{lem:finite_sample_joint_ci}  are valid for all $n \in \NN$, they may be conservative in practice. To overcome this, we construct asymptotic confidence intervals based on efficiency theory and covariance-adjusted multiple testing corrections below.

\subsection{Asymptotic Joint Lower Bounds} Our asymptotic confidence intervals are based on the doubly-robust estimation approaches from the causal inference literature \cite{dml, athey_wager_policy_learning, uehara2022reviewoffpolicyevaluationreinforcement}. We start by splitting $\{O_i\}_{i=1}^n$ randomly into $L$-folds: $D_1,...,D_F$, and denote $f(i)$ as the fold containing the $i$-th observation. We estimate $\{\hat{\mu}^f\}_{f=1}^F$ as the estimated conditional means  trained on all folds except the $f$-th fold.\footnote{Note that $\hat{\mu}$ and $\hat{e}$ may be estimated with nonparametric estimators, such as neural networks or random forests.} After the cross fitting procedure, we estimate policy values $V_j(\pi)$ using Definition \ref{defn:dml}, and conduct inference based on asymptotic normality results in Lemma \ref{lem:dml_normal}.

\begin{definition}[Efficient Policy Value Estimation] \label{defn:dml}
For $\pi \in \widetilde\Pi$ for a fixed policy set $\widetilde{\Pi}$, we estimate ${V}_j(\pi)$, policy $\pi$'s $j$th policy value, as 
\begin{equation}
        \hat{V}_j^{DR}(\pi) =\sum_{i=1}^n \psi_j(O_i, \pi; \hat\mu^{f(i)}, {e}),
\end{equation}
where  $\psi_{jk}^{DR}(O; \mu, e) = \frac{\mathbf{1}[A=k]\left(Y_j - \mu_j(k, X) \right)}{e(k,X)} + \mu_j(k,X),$ and $\psi_{j}^{DR}(O, \pi; \mu, e) = \sum_{k \in [K]}  \pi(k, X)   \psi_{jk}(O; \mu, e).$
\end{definition}

The additional term in $\psi_{jk}$ as opposed to $\psi_{jk}^{IPW}$ is a regression term that reduces the variance of our estimates, and enables $\hat{V}_j(\pi)$ to achieve the smallest possible variance among all possible estimators under minimal conditions\footnote{To be precise, the estimator in Definition \ref{defn:dml} achieves the smallest possible asymptotic variance in a class of regular and asymptotically linear estimators. For further details, we refer to \cite{Vaart_1998}.}. Definition \ref{defn:joint_asymp_lower_bds} provides our asymptotic joint lower confidence bounds.

\begin{definition}[Joint Asymptotic Lower Confidence Bounds]\label{defn:joint_asymp_lower_bds}
    Without loss of generality, assume that $\Scal = \{1,..., |\Scal|\}$. Let $\widetilde{\Pi}= \{\pi_i\}_{i=1}^{T}$ denote a fixed set of policies, and let $\Scal$ be the set of guardrail outcomes. Let $\hat{D}_j(\pi) = \hat{V}_j^{DR}(\pi) - (1-w_j)\hat{V}_j^{DR}(\pi_0)$, and let $\hat{d}_j(O_i, \pi) = \psi_j^{DR}\left(O_i, \pi, \hat{\mu}_{f(i)}, {e}\right) - \left(1+w_j\right)\psi_j^{DR}\left(O_i, \pi_0, \hat{\mu}_{f(i)}, {e}\right) $. Denote the $(|\Scal|\times |\widetilde{\Pi}|) \times (|\Scal|\times|\widetilde{\Pi}|)$-empirical covariance matrix $\widehat{\Sigma}$, with the $\left((p_1-1)|\Scal| + s_1, (p_2-1)|\Scal|+s_2\right)-$th entry defined as 
    $$ \frac{1}{n}\sum_{i=1}^n \left( \hat{d}_{s_1}(O_i, \pi_{p_1}) - \hat{D}_{s_1}(\pi_{p_1})  \right)\left( \hat{d}_{s_2}(O_i, \pi_{p_2}) - \hat{D}_{s_2}(\pi_{p_2}) \right)  $$
    for all $s_1, s_2 \in \{1,\dots, |\Scal|\}$ and $p_1, p_2 \in \{1,\dots,T\}$.

    Define $z^*_{n_{\text{sim}}}(\alpha)$ as the lower $\alpha$-quantile of $\min_{j} \widehat\Sigma_{jj}^{-1/2}\tilde\rho_j$ over $n_{\text{sim}}$ simulated i.i.d. multivariate normal vectors $\tilde\rho \sim N(\mathbf{0}, \widehat{\Sigma})$. We denote our asymptotic lower bound confidence sets for policies $\pi_k \in \widetilde{\Pi}$ as $\hat{\Ccal}^{DR}(\pi_k, \alpha)$, where the $j$th lower bound in $\hat{C}^{DR}(\pi_k, \alpha)$ is defined as 
    \begin{equation}
        \hat{C}_j^{DR}(\pi_k, \alpha) = \left(\hat{D}_j(\pi) - \frac{z^*_{n_\text{sim}}(\alpha)}{\sqrt{n}} \widehat{\Sigma}_{\left((k-1)|\Scal|+j, (k-1)|\Scal|+j \right)}\right).
    \end{equation}
\end{definition}

Our lower confidence bound construction differs from the finite sample lower bound construction in Lemma \ref{lem:finite_sample_joint_ci} in two distinct ways: (i) it leverages an asymptotically normal approximation instead of a finite-sample Bernstein inequality, and (ii) uses a multiple testing correction based on the covariances across each $\pi \in \widetilde{\Pi}$ and guardrail outcomes $j \in \Scal$. In particular, our procedure uses the sup-$t$ bound correction, which has been shown to be less conservative than union-bound based corrections both theoretically \cite{sup_t_band} and in practice \cite{cho2024cspimtcalibratedsafepolicy}. This correction sets the critical value as the smallest uniform lower confidence bounds for $V_j(\pi) - (1+w_j)V_j(\pi_0)$ such that the lower bound widths are proportional to the estimated standard errors. Definition \ref{defn:joint_asymp_lower_bds} computes the adjusted critical score $z_{n_\text{sim}}^*(\alpha)$ by randomly sampling $n_\text{sim}$ multivariate zero-mean normal vectors with the empirical covariance matrix $\widehat{\Sigma}$.

Lemma \ref{lem:asymp_err_control} provides the sufficient conditions, which match standard regularity assumptions for asymptotic error control in the doubly-robust estimation literature \cite{dml}.

\begin{lemma}[Asymptotic Validity of Definition \ref{defn:asymp_safe_alg}]\label{lem:asymp_err_control}  Assume that (i) estimated functions $\hat{\mu}_l$, $\hat{e}_l$ are bounded with respect to $P$ almost surely, (ii) $\|\hat{\mu}_l - \mu \|_{P, 2}  = o_P(1)$, and (iii) $\text{Var}(Y|A=a,X=x)$ be bounded below for all $a \in [K], X=x$.

Then, as $n \rightarrow \infty$, for $n_{\text{sim}}\rightarrow \infty$, for all $\pi \in \Pi$, $j \in \Scal$, $\hat{C}_j^{DR}(\pi, \alpha)$ are joint lower confidence bounds for $V_j(\pi) - (1+w_j)V_j(\pi_0)$, i.e.,
    \begin{align}
        \limsup_{n \rightarrow \infty} \lim_{n_\text{sim}\rightarrow \infty} &P\biggr{(}\exists \pi \in \widetilde{\Pi}, \ j \in \Scal: \\
        & \left(V_j(\pi) - (1+w_j)V_j(\pi_0) < \hat{C}_j^{DR}(\pi, \alpha)\right) \biggr{)} \leq \alpha. 
    \end{align}
\end{lemma}

Given our method of constructing joint lower confidence bounds for policy values over a fixed policy set $\widetilde{\Pi}$, we now turn to constructing lower bounds with the same guarantees for data-adaptively selected policy sets $\widehat{\Pi}$. To do so, we first define the notion of $\epsilon$-stability from the differential privacy literature.

\subsection{Algorithmic Stability and Post-Selection Correction}\label{sec:4_3}

Intuitively, the post-selection effects for inference on policy set $\widehat\Pi$ will lessen if the learned policy set is relatively robust to the observed data $\{O_i\}_{i=1}^n$. We formalize this notion with concept of $\epsilon$-stability \cite{bassily2015algorithmicstabilityadaptivedata}, which quantifies the sensitivity of policy selection.

\begin{definition}[$\epsilon$-stability \cite{bassily2015algorithmicstabilityadaptivedata}]\label{defn:stable}
Let $\Acal: \{O_i\}_{i=1}^n \rightarrow \Pi$ be a randomized algorithm from the observed data. We say that an algorithm $\Acal$ is $\epsilon$-stable, if for every pair of $n$-tuples $\{O_i\}_{i=1}^n$ and $\{O_i'\}_{i=1}^n$ that differ by one observation $O_i$, and any measurable set $\Ecal$, 
\begin{equation}
    \PP\left(\Acal\left( \{O_i\}_{i=1}^n\}\right) \in \Ecal \right) \leq e^{\epsilon} \PP\left( \Acal\left( \{O_i'\}_{i=1}^n\}\right) \in \Ecal  \right),
\end{equation}
where measure $\PP$ is taken only over the randomness of the algorithm. 
\end{definition}

Because the event of miscoverage (i.e. confidence intervals not containing the true $\tau_j(\pi)$ for all $j \in \Scal$) is measurable, this implies the following correction for our confidence intervals.

\begin{figure}
    \centering
    \includegraphics[width=0.5\linewidth]{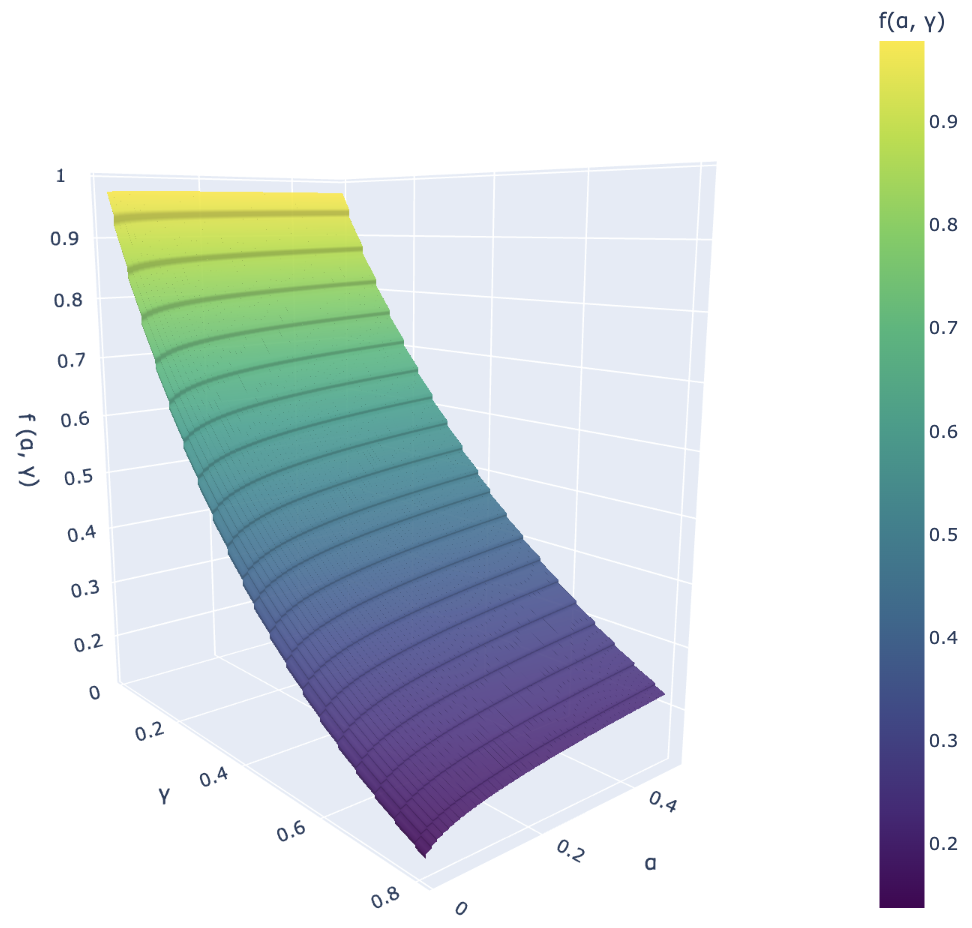}\Description{}
    \caption{Plot of $f(\gamma, \alpha)$ for $\alpha \in [0.01, 0.5]$, $\gamma \in [0.01, 0.8]$. }
    \label{fig:alpha_gamma_plot}
\end{figure}

\begin{lemma}[Post-Selection Correction for $\epsilon$-stable Policy Selection]\label{lem:post_selection_correction}
    Let $\alpha'(\delta) = (\alpha - \delta)\exp(-\frac{n}{2}\epsilon^2 -\epsilon\sqrt{n\log(2/\delta)/2})$. 
    If $\widehat\Pi$ is the output of an $\epsilon$-stable algorithm, then for all $\delta \in (0, \alpha)$,  $\hat{C}^{IPW}\left(\pi, \alpha'(\delta)\right)$ as defined in Lemma \ref{lem:finite_sample_joint_ci} provides $\alpha$-level uniform error guarantees across all $\pi \in \widehat{\Pi}$, $j \in \Scal$ for all $n \in \NN$, i.e., 
    \begin{align}
      \forall n \in \NN, \   P\biggr{(}\exists \pi \in \widehat{\Pi}, \ j \in \Scal &: (V_j(\pi) - \left(1+w_j\right)V_j(\pi_0)  ) \\ 
        &\leq  \hat{\Ccal}_j^{IPW}\left(\pi, \alpha'(\delta)\right) \biggr{)} \leq \alpha.
    \end{align}
    Furthermore, for all $\delta \in (0, \alpha)$, under the same assumptions as Lemma \ref{lem:asymp_err_control}, $\hat{C}^{DR}\left(\pi, \alpha'(\alpha)\right)$ provides $\alpha$-level uniform error guarantees across all $\pi \in \widehat{\Pi}$, $j\in\Scal$ for $n_\text{sim} \rightarrow \infty$ as $n \rightarrow \infty$, i.e, 
    \begin{align}
        \limsup_{n \rightarrow \infty} \lim_{n_\text{sim}\rightarrow \infty} &P\biggr{(}\exists \pi \in \widehat{\Pi}, \ j \in \Scal: \\
        & \left(V_j(\pi) - (1+w_j)V_j(\pi_0) < \hat{C}_j^{DR}\left(\pi, \alpha'(\delta)\right)\right) \biggr{)} \leq \alpha. 
    \end{align}
\end{lemma}
We obtain these results with a direct application of Definition~\ref{defn:stable}. To ensure $\alpha'(\delta)$ remains a constant for a fixed $\alpha \in (0,1)$ and $\delta \in (0, \alpha)$ as $n \rightarrow \infty$, $\epsilon$ can be set as $\gamma/\sqrt{n}$, where $\gamma$ is a fixed constant. The parameter $\delta$ is a free variable, and thus a natural choice of $\delta$ is to pick $\delta \in (0,\alpha)$ that maximizes $\gamma$. In Figure \ref{fig:alpha_gamma_plot}, we show a plot of $f(\gamma, \alpha) = \argmax_{\delta \in (0, \alpha)}\alpha'(\delta) / \alpha$ as a function of $\gamma  =\epsilon \sqrt{n}$ and $\alpha$. The results show that setting $\gamma$ as a small constant keeps $\alpha'(\delta)$ close to $\alpha$, while large values of $\gamma$ result in rapid decay of the corrected error level $\alpha'(\delta)$.  

Lemma \ref{lem:post_selection_correction} provides a valid method for constructing valid joint lower confidence bounds for a policy set $\widehat{\Pi}$ selected by $\epsilon$-stable algorithm using the entire dataset $\{O_i\}_{i=1}^n$. In the following section, we introduce our final algorithm, \emph{safe policy noisy learning} (SNPL), which combines the results of Lemma \ref{lem:post_selection_correction} and an $\epsilon$-stable policy selection approach in order to perform safe policy learning without requiring (i) data-splitting or (ii) joint inference over all $j \in \Scal$ for each $\pi \in \Pi$, the entire candidate policy set.

\section{Safe Policy Learning on Offline Data} \label{sec:p2}

At a high level, SNPL prunes the candidate policy class $\Pi$ to a smaller set $\widehat{\Pi}$ in a data-driven, $\epsilon$-stable manner. Following this pruning stage, our approach constructs joint lower confidence bounds for $\widehat{\Pi}$ using the corrections in Lemma \ref{lem:post_selection_correction}. The final policy $\hat\pi$ is the policy with the largest estimated goal value among the policies whose post-correction lower bounds are all above zero, indicating that the policy satisfies the desired policy changes with high probability. If no selected policies satisfy these criteria, then our algorithm returns the baseline policy, ensuring its safety.

The full pseudocode for our approach is presented in Algorithm~\ref{alg:safe_policy_learning}, which takes in a generic policy value estimator $\hat{V}$ (for selecting a final policy) and lower confidence bound method $\hat{C}$ (for ensuring safety).\footnote{We clarify possible confusions in our pseudocode in Algorithm \ref{alg:safe_policy_learning}\label{footnote:alg1}. In line \ref{line:4}, lower bounds may possibly depend on the final pruned set $\widehat{\Pi}$, which is unknown until our pruning process terminates. We provide additional details in Appendix \ref{sec:appendix_exp}.}. With the correct choice of hyperparameters, we achieve $(\Scal, \alpha, w)$-safety with $\hat{V}_g^{IPW}$ and $\hat{C}^{IPW}$ in Definition \ref{defn:ipw} and  Lemma \ref{lem:finite_sample_joint_ci} respectively. Likewise, we achieve  $(\Scal, \alpha, w)$-asymptotic safety with $\hat{V}_g^{DR}$ and $\hat{C}^{DR}$ in Definitions \ref{defn:dml} and \ref{defn:joint_asymp_lower_bds}. We provide both theoretical results and heuristics for hyperparameter selection below.

\paragraph{Ensuring $\epsilon$-stability} Based on Lemmas \ref{lem:finite_sample_joint_ci} and \ref{lem:asymp_err_control}, a policy $\pi$ satisfies the desired changes in guardrail policy values with high probability as long as all adjusted joint lower confidence bounds are above zero. While our algorithm would ideally only include $\pi$ in $\widehat{\Pi}$ if its minimum lower bound is above zero, this approach is deterministic for a fixed dataset, fails to be $\epsilon$-stable, and does not permit post-selection corrections using Lemma \ref{lem:post_selection_correction}. 

To maintain $\epsilon$-stability, our algorithm takes in three distinct parameters: a sensitivity parameter $B$, a noise parameter $\epsilon$, and the maximum number of selected policies $\eta$. For each policy $\pi \in \Pi$, instead of directly thresholding the minimum lower bound on zero, we use noisy versions of the estimated lower bound and the zero threshold by adding independent noise generated from a Laplace distribution. The parameters $B$ and $\epsilon$ determine the scale of the noise injected. When either all policies have been scanned or $\eta$ number of policies have been selected, we conclude our pruning procedure and return $\widehat{\Pi}$ for the selection process described above. We establish the safety guarantees of Algorithm \ref{alg:safe_policy_learning} for the correct choice of $B$, formalized in Theorem \ref{thm:safe_policy_search} below.



\begin{algorithm}[t!] 
  \caption{Noisy Safe Policy Learning (NSPL)} \label{alg:safe_policy_learning}
  \begin{algorithmic}[1] 
  \State 
  \textbf{input}: data 
  $\{O_i\}_{i=1}^n$, $g$, $\Scal$, $w$, error level $\alpha$, policy class $\Pi$, propensity score bound $c$, baseline policy $\pi_0$, sensitivity parameter $B$, stability parameter $\epsilon$, maximum search size $\eta \in \big{[}|\Pi|\big{]}$, lower confidence bound construction method $\hat{C}$, estimator $\hat{V}_g$. 

  \State Set $\widehat\Pi$ as the empty set, and $\delta^* = \argmax_{\delta \in (0,\alpha)} \alpha'(\delta)$, where $\alpha'(\delta)$ is as defined in Lemma \ref{lem:post_selection_correction}.\label{line:2}

  \State Sample a threshold value $v \sim \text{Lap}(2B\eta/\epsilon)$. \label{line:3}

  \For{$\pi \in \Pi \setminus \{\pi_0\}$}

  \State For $j \in \Scal$, estimate the joint lower bounds $\hat{C}\left(\pi, \alpha'(\delta^*)\right)$.\label{line:4}

  \State Set $M'(\pi) = \min_{j \in \Scal} \hat{C}_j\left(\pi, \alpha'(\delta^*)\right)$.

  \State Generate independent noise $\epsilon_\pi \sim \text{Lap}(4B\eta/\epsilon)$. \label{line:7}

  \If{${M}'(\pi) + \epsilon_\pi > v$} set $\widehat\Pi = \widehat\Pi\cup \{\pi\}$.
  \EndIf
  \If{$|\widehat\Pi| = \eta$} 
  \textbf{break}.
  \EndIf
  \EndFor

  \State For each $\pi \in \widehat{\Pi}$, construct joint lower bounds $\hat{C}\left(\pi, \alpha'(\delta)\right)$.

  \State Set $\widehat{M}(\pi) = \min_{j \in \Scal} \hat{C}_j(\pi, \alpha)$ for each $\pi \in \widehat{\Pi}$. \label{algline:line15} 

  \If{$\{\pi \in \widehat\Pi: \widehat{M}(\pi) > 0 \}$ is nonempty} set $\hat\pi = \argmax_{\pi \in \widehat\Pi: \widehat{M}(\pi) > 0 } \hat{V}_g(\pi)$. 
  \Else{ \ set  $\hat\pi = \pi_0$.}
  \EndIf

  \State \textbf{Return} $\hat{\pi}$.
\label{alg:spl}
  \end{algorithmic}    
\end{algorithm}




\begin{theorem}[Asymptotic Safety of Algorithm \ref{alg:safe_policy_learning}]\label{thm:safe_policy_search} 
Let $\xi = (2+\max_{j \in \Scal} w_j)/c$, and let $B_{\text{finite}} = \frac{2\xi}{n}$ + $\sqrt{2\log\left(\frac{3}{\alpha'(\delta)}\right) t(n,\xi)}$, where 
\begin{equation}
    t(n,\xi) = \frac{1}{n(n-1)}\left(4\xi^2 + \frac{2\xi^2}{n} + \frac{2\xi^2(n-1)}{n}\right).
\end{equation}

Consider Algorithm \ref{alg:safe_policy_learning} using the confidence bound method $\hat{C}^{IPW}$ defined in Lemma \ref{lem:finite_sample_joint_ci}. For all $B \geq B_{\text{finite}}$, $\epsilon > 0$, $\eta \in [|\Pi|]$, the selection of $\widehat{\Pi}$ is $\epsilon$-stable, and Algorithm \ref{alg:safe_policy_learning} is $(\Scal, \alpha, w)$-safe.

Likewise, let $B_{\text{asymp.}} = \frac{4\xi}{n} + \Phi^{-1}\left(1-\frac{\alpha'(\delta^*)}{\eta |\Scal|}\right)\sqrt{t(n,2\xi)}$. Under the same conditions as Lemma \ref{lem:asymp_err_control}, for all $B \geq B_{\text{asymp.}}$, $\epsilon > 0$, $\eta \in [|\Pi|]$, the selection of $\widehat{\Pi}$ is $\epsilon$-stable, and Algorithm \ref{alg:safe_policy_learning} using the confidence bound method $\hat{C}^{DR}$ in Definition \ref{defn:joint_asymp_lower_bds} is $(\Scal, \alpha, w)$-safe asymptotically. 
\end{theorem}

Theorem \ref{thm:safe_policy_search} provides both theoretical and practical guidelines for setting $B$, one of the key parameters for constructing policy set $\widehat\Pi$. As $B$ grows large, the noise added to both the minimum lower bound estimates $\widehat{M}(\pi)$ and the ideal zero grows larger in scale. Therefore, $B$ should be set as small as possible to avoid an unnecessarily noisy policy search, while still maintaining the desired safety guarantees. Theorem \ref{thm:safe_policy_search} provides $B_\text{finite}$ and $B_\text{asymp.}$ as natural choices for Algorithm \ref{alg:safe_policy_learning}.

\paragraph{Heuristics for Hyperparameter Selection} Importantly, our choice of $B_{\text{finite}}$ (or $B_{\text{asymp.}}$) and $\epsilon = \gamma / \sqrt{n}$ (discussed in Section \ref{sec:4_3}) ensures that the added noise converges to zero as $n$ grows large. Both $B_\text{finite}$ and $B_\text{asymp.}$ are on the order of $1/n$, implying that $B/\epsilon$ is on the order of $1/\sqrt{n}$. Therefore, the scale of the added Laplacian noise in Lines \ref{line:3} and \ref{line:7} converges to zero as $n \rightarrow \infty$ as long as $\eta$ and $\gamma$ are fixed constant with respect to the sample size $n$. In practice, this implies that with large datasets, our pruning procedure will mostly include policies that satisfy the desired changes. 

\textbf{Selection of $\gamma$.}  We base our selection of $\gamma$ based on the ratio of $\alpha'(\delta^*)$ (as defined in Line \ref{line:2} of Algorithm \ref{alg:safe_policy_learning}) to the desired error level $\alpha$. As the ratio $\alpha'(\delta^*)/\alpha$ grows small, our lower confidence bounds grow wider in magnitude, leading to difficulties in finding valid policies $\pi$ that satisfy our safety criteria in Line \ref{algline:line15} of Algorithm \ref{alg:safe_policy_learning} Based on Figure \ref{fig:alpha_gamma_plot}, we see that setting $\gamma < 0.5$ allows us to maintain at least a ratio of roughly $1/2$ across different $\alpha$-values. 

\textbf{Selection of $\eta$. } We propose a heuristic guideline based on comparing lower confidence bound widths between our finite-sample joint lower bounds used in Algorithm \ref{alg:safe_policy_learning} and the finite-sample joint lower bounds in \ref{lem:finite_sample_joint_ci} with a Bonferroni correction \cite{Dunn_1961} over the entire policy class. As $n\rightarrow \infty$, the joint lower confidence bound widths post Bonferroni correction over the entire policy class $\Pi$ scale roughly at a factor of $\log\left(\frac{|\Scal| |\Pi|}{\alpha}\right)$. In contrast, our adjusted confidence bounds have widths that scale by a factor of at most $\log\left(\frac{\eta |\Scal|}{\alpha'(\delta^*)}\right)$. To obtain lower bound widths that scale as at most a $p$-fraction of the rate obtained by the Bonferroni correction, our results suggest that we require the max size $\eta \leq {\alpha\left(\delta^*\right)}{\frac{|\Pi|^p}{\alpha^p|\Scal|^{1-p}}}$, where $p < 1$. Standard choices of $p$ are $p=2/3$ or $p=1/2$, ensuring that our lower bound widths shrink at roughly 1.5 or 2 times the rate of Bonferroni-corrected bounds as $n \rightarrow \infty$.

\begin{table}[t!]
    \centering\caption{Heuristics for Hyperparameter Selection.}
    \label{tab:hypereparameters}
\begin{tabular}{c|c}
Parameter & Heuristic \\ \hline
$B$ & $B_\text{finite}$ or $B_\text{asymp}$ in Theorem \ref{thm:safe_policy_search}.\\ \hline 
 $\epsilon$ & $\frac{\gamma}{\sqrt{n}}$ for $\gamma \in (0, 0.5]$. \\ \hline 
 $\eta$ & $  \max\left({\frac{\alpha'(\delta^*)|\Pi|^p}{\alpha^p|\Scal|^{1-p}}}, 1\right)$ for $p < 1$.\\ 
\end{tabular}
\vspace{-1em}
\end{table}
To summarize our heuristics for  hyperparameter selection, we provide our guidelines on $B$, $\epsilon$, and $\eta$ in Table \ref{tab:hypereparameters} above.

\section{Empirical Results}\label{sec:experiments}

\begin{figure*}[h]
    \centering
    \includegraphics[width=0.45\linewidth]{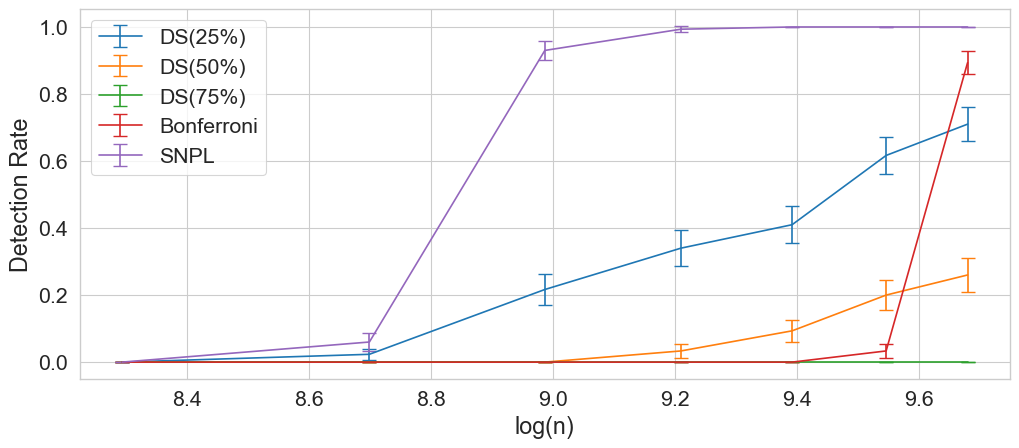}
    \includegraphics[width=0.45\linewidth]{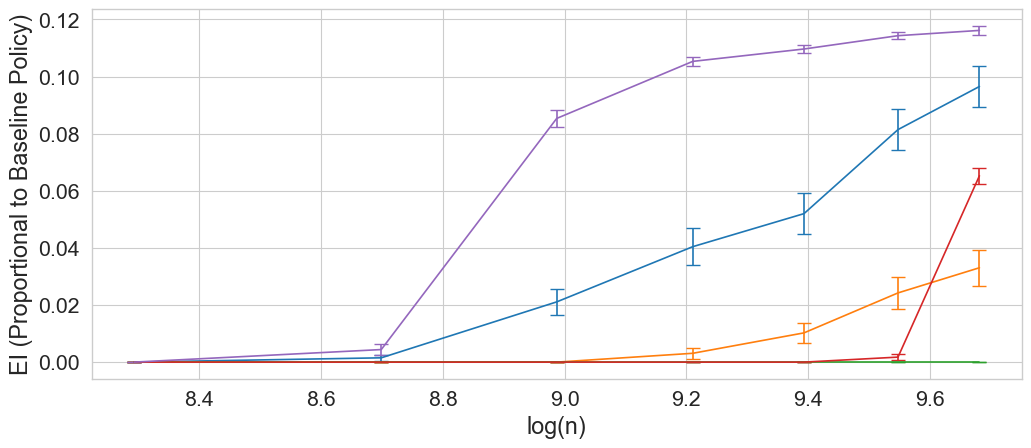}
    \includegraphics[width=0.45\linewidth]{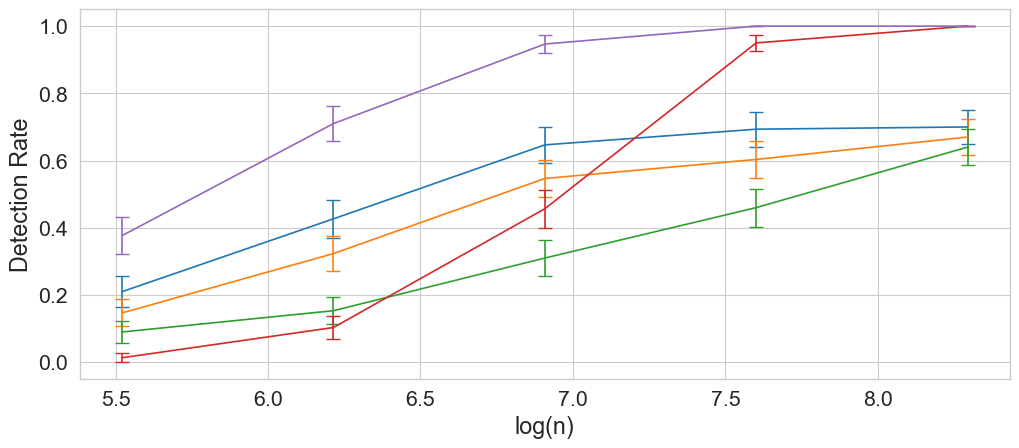}
    \includegraphics[width=0.45\linewidth]{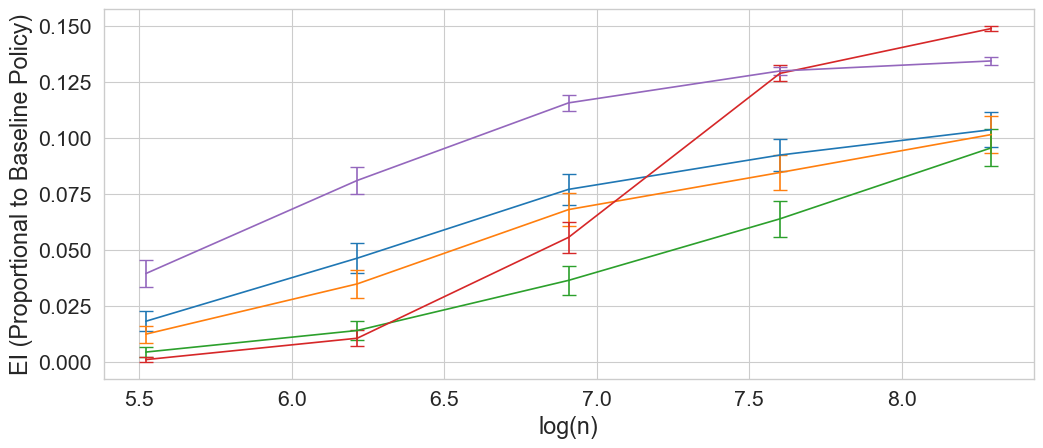}
    \caption{Detection rates and EI on our real-world case study. The top row corresponds to performance with finite-sample safety guarantees, while the bottom row corresponds to performance with asymptotic guarantees.}
    \label{fig:main_plot}
\end{figure*}

This section presents results for a real-world case study from a large consumer technology company. In Appendix \ref{sec:appendix_exp}, we provide further details, including pseudocode for our baselines and additional synthetic experiments. 

\paragraph{Dataset and Policies} The dataset is collected from a randomized control trial with a fixed propensity score of $1/2$. The experiment tested $K=2$ treatments, with two outcomes regarding SMS delivery cost and user retention levels. Although the complete dataset comprises approximately 1.2 million observations, the subset reserved for policy selection and evaluation (i.e., test dataset) contains roughly 13,000 observations.\footnote{The test set is relatively small compared to the whole dataset due to most of the data being used to construct the policy class, which roughly forms a pareto frontier for the two outcomes. Because this data is used to learn the policy class itself, it cannot be used for safe policy learning.} Our policy class $\Pi$ consists of 2949 contextual policies, parameterized by the linear threshold policy class $\Pi$ \cite[Equation 1]{garrard2023practicalpolicyoptimizationpersonalized}\footnote{Note that these policies directly leverage information from treatment effect estimators trained with data excluding the test set. As a result, policy selection and evaluation are strictly done on the test portion of the data to ensure statistical validity.}.
Notably, the dataset examined in this study features policies with remarkably low signal-to-noise (SNR) values.
We define SNR as the ratio of the estimated gap between a policy's $j$th value against its corresponding guardrail constraint, divided by the standard error of its estimate. In the test data, the SNR ratio over both guardrail constraints and all candidate policies can be as small as $\sim0.008$, even with 13,000 observations.

\paragraph{Metrics for Safe Policy Improvement} For safe policy improvement, we focus on three specific metrics: (i) detection rate, (ii) type I error, and (iii) expected improvement (EI). We define detection rate as the probability that an algorithm returns a policy different from baseline. For Type I error, we calculate the probability that a returned policy satisfies our constraints defined by weights $w$ on \emph{all} guardrail outcomes, conditional on the returned policy differing from the baseline. EI measures the average gain in the goal metric for the returned policy $\hat\pi$. To  obtain these averages and probabilities, we utilize the true values (known for synthetic cases and estimated for the real-world example) across 300 simulations, with error bars corresponding to 2 standard deviations.

\begin{figure*}[h]
    \centering
    \includegraphics[width=0.47\linewidth]{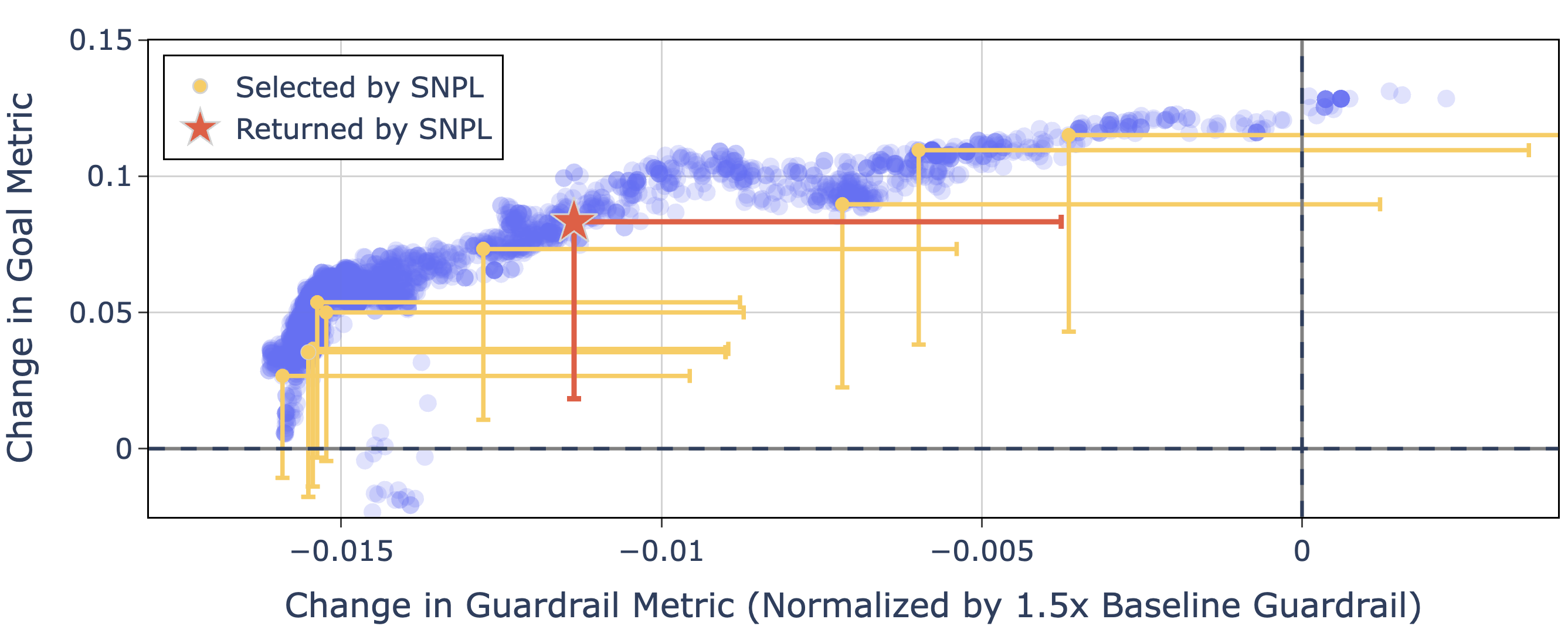}\Description{}
    \includegraphics[width=0.46\linewidth]{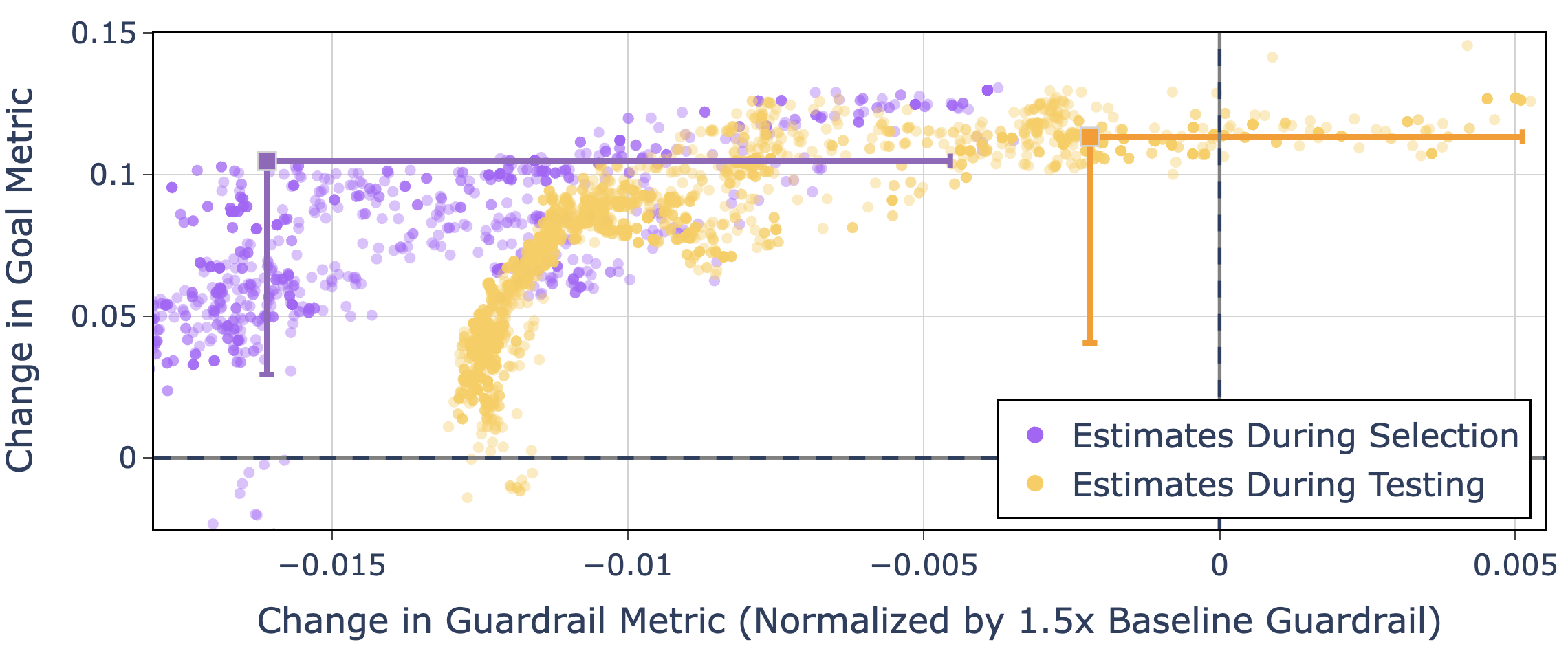}\Description{}
    \caption{Visualization of SNPL (left) and Data-Splitting with 50/50 Split (right) for $n=500$ using asymptotic approaches.}
    \label{fig:story_plot}
\end{figure*}

\paragraph{Experiment Settings} For all experiments, we set the error level $\alpha = 0.1$. Both synthetic and real-world experiments employ the same parameter settings for Algorithm \ref{alg:safe_policy_learning}. Specifically, our $(\Scal, \alpha, w)$-safe variant of the algorithm utilizes $\hat{V}_g$ as defined in Definition \ref{defn:ipw}, $\hat{C}$ from Lemma \ref{lem:finite_sample_joint_ci}, and $B = B_\text{finite}$ for Algorithm \ref{alg:safe_policy_learning}. For our $(\Scal, \alpha, w)$-asymptotically safe variant, we set $B = B_\text{asymp.}$ and use $\hat{V}_g, \hat{C}$ in Definitions \ref{defn:dml} and \ref{defn:joint_asymp_lower_bds}, respectively. Across both variants of Algorithm \ref{alg:safe_policy_learning}, we set $\gamma = 0.1$ (i.e, $\epsilon = \gamma/\sqrt{n}$) and $\eta = 10$, following the heuristics in Table \ref{tab:hypereparameters} with $p = 0.5$. We set the baseline as giving the entire population treatment $A=1$. To provide true policy values $V(\pi)$ for each outcome, we use estimated policy values obtained with $\hat{V}^{DR}_g$ in Definition \ref{defn:dml} using the entire dataset. We set our goal outcome to maximize gains in user retention (goal outcome), while setting both SMS delivery cost and user retention as guardrail outcomes. While we optimize for user retention as our goal outcome, we additionally include user retention as a guardrail to ensure probabilistic guarantees of improvement. We set $w = [0, -1/2]$, indicating that acceptable policies will only improve (increase) user retention, while allowing for up to 50\% increase in cost (i.e., 50\% decrease in cost reduction)\footnote{While the main body of our text assumes that we want to maintain large values for guardrail metrics, our application aims to ensure that one of the guardrails, SMS cost, does not exceed a certain threshold. To accommodate this, we simply use upper bounds, rather than lower bounds, for the guardrail metric of SMS costs. Upper bounds can directly be obtained from our lower confidence bounds in Section \ref{sec:p1} by adding, rather than subtracting, the term with standard errors.}. To obtain different signal-to-noise regimes, we vary $n$, the total number of samples, by sampling with replacement from the 13,000 samples in the test dataset. For our $(\Scal, \alpha, w)$-safe approach, we test $n$ from 4000 to 16000 in increments of 2000 samples. For our $(\Scal, \alpha, w)$-asymptotically safe approach, we set $n \in \{250, 500, 1000, 2000, 4000\}$. The tested ranges differ due to trivial results for $(\Scal, \alpha, w)$-safe approaches with small $n$ (and vice versa for our asymptotic approach), shown in Figure \ref{fig:main_plot}. 


\paragraph{Baseline Approaches} 
As baselines for our approach, we employ three data-splitting approaches (DS) and a Bonferroni correction over all candidate policies $\Pi$.  To evaluate the effectiveness of data-splitting, we adapt the CSPI/HCPI method \cite{cho2024cspimtcalibratedsafepolicy, pmlr-v37-thomas15} that selects the policy on 25\%, 50\% and 75\% of the data, and evaluates the learned policy on the complement data portion. Specifically, we estimate confidence lower bounds (as described in Section \ref{sec:p1}) on the complement data. To ensure safety, we only return a policy if its minimum lower confidence bound constructed on the complement portion is above zero. For the Bonferroni approach, we use the Bonferroni correction \cite{Dunn_1961} for all guardrail policy values over all policies $\pi \in \Pi$. We select the policy that maximizes the estimated improvement in the goal metric among those policies whose corrected minimum lower bounds are greater than zero. To obtain $(\Scal, \alpha, w)$-safety guarantees, we use the lower confidence bounds $\hat{C}^{IPW}$ in Lemma \ref{lem:finite_sample_joint_ci} with estimator $\hat{V}_g^{IPW}$ in Definition \ref{defn:ipw}. To obtain $(\Scal, \alpha, w)$-asymptotic safety guarantees, we use $\hat{C}^{DR}$ in Definition \ref{defn:joint_asymp_lower_bds} and $\hat{V}_g^{DR}$ in Definition \ref{defn:dml}. We refer to the discussion in Section \ref{disc:baselines} for why other safe policy learning approaches are not applicable for our setting. For further details on our baselines, we refer to Appendix \ref{sec:appendix_exp}.

\paragraph{Discussion of Empirical Results}

Our empirical results demonstrate the benefits of SNPL on real-world data. Figure \ref{fig:main_plot} plots detection rates and EI as a function of $n$ across both $(\Scal, \alpha, w)$-safe and asymptotically safe approaches. Type I error is stable across all tested methods. Both SNPL and Bonferroni approaches report zero Type-I error across all $n$ for both types of safety guarantees. DS approaches report Type I error less than $\alpha=0.1$ for $n > 500$ with asymptotic guarantees, and Type 1 error of zero with finite-sample guarantees. For asymptotic DS with $n\leq 500$, the largest reported Type I error is 0.44 for DS (75\%). This is to be expected with small $n$ using asymptotic guarantees. 

For all values of $n$, SNPL outperforms all baselines in terms of detection rate across both $(\Scal, \alpha, w)$-safe and asymptotically safe variants. As noted above, SNPL's improved detection rate does not come at the cost of higher Type I error, as SNPL has zero Type I error across all experiments. As $n$ grows, the detection rate across all approaches grows larger, with SNPL uniformly achieving the largest detection rate for both finite-sample and asymptotic variants. SNPL provides the largest improvements for detection rate at intermediate values of $n$ across the ranges tested for asymptotic and nonasymptotic safety guarantees. With asymptotic guarantees, SNPL improves detection rates by roughly 0.3 for $n=1000$, resulting in a 50\% increase in detection rate relative to the best-performing baseline. With nonasymptotic guarantees, SNPL increases detection rate by more than 0.6 for $n=8000$, providing a 300\% increase in relative performance compared to the best-performing baseline. 

SNPL boasts larger EI when SNR remains relatively small\footnote{We explore synthetic examples with larger SNR regimes to verify this. We refer readers to Appendix \ref{sec:appendix_exp} for these results}. As $n$ increases, note that the SNR ratio increases on the order of $\sqrt{n}$. When the SNR grows large, estimates are more stable across splits, and lower bounds for all approaches converge to the estimated policy values. The Bonferroni baseline in particular benefits from this, as it is able to select among \emph{all} candidate policies, rather than a random subset, as confidence lower bounds grow small. As a result, Bonferroni outperforms SNPL at $n=4000$ for asymptotically safe approaches. For approaches that guarantee safety for all $n$, we note that even $n=16000$ is not large enough for this to occur. 

Our experiments demonstrate the value of SNPL in practical settings. At roughly the true test set sample size ($n=12000)$, SNPL with nonasymptotic safety guarantees boasts detection rates of near 1 (0.993), while all other baselines at best have detection rates of 0.34. As a result, the EI achieved by SNPL improve upon the best performing baselines's EI by 150\%. 

To explore why SNPL provides drastic improvements over other baselines, in Figure \ref{fig:story_plot} we visualize the estimated lower bounds, normalized according to weight vector $w = [0, -0.5]$,  for one run of our asymptotic approach with $n=500$. To be deemed safe, the bounds for the selected policy must be within the upper-left quadrant delineated by the black vertical and horizontal lines. In low SNR regimes, estimates of policy changes shift between different data splits, as shown in the right panel. As a result, the policy selected by DS, which is estimated to be safe during selection (shown in purple) is not deemed as safe when tested on the data complement (shown in yellow). In contrast, by avoiding data-splitting, SNPL is able to freely pick among its pruned policy set $\widehat{\Pi}$ (shown in yellow in the left panel) without additional randomness in its final selection, resulting in a safe policy choice indicated by the dark orange star. We note that the confidence bounds with Bonferroni corrections all cross zero on at least one axis in this synthetic run. By using data-driven pruning of $\Pi$, SNPL effectively enables safe policy learning when (i) data-splitting approaches suffer from instability and (ii) Bonferroni corrections result in weak statistical power.


\section{Conclusions}
This work introduces safe noisy policy learning (SNPL), a method for offline policy learning intended for practical settings where SNR is low and policy classes are large. SNPL enables simultaneous policy learning and inferences, avoiding the inefficiencies of data-splitting or multiple testing corrections that occur in this setting. Real-world experiments demonstrate that SNPL improves the detection of safe policies, maintains safety error levels, and achieves higher expected goal outcomes compared to alternative methods with similar theoretical guarantees.

\newpage
\bibliographystyle{plain}
\balance
\bibliography{citation}

\appendix

\section{Proofs} \label{sec:appendix_proofs}
In this section, we provide simple proofs for all theoretical results presented in our work.

\paragraph{Proof of Lemma \ref{lem:finite_sample_joint_ci}}\label{proof:lem:finite_sample_joint_ci} To prove this result, we leveraging an existing empirical Bernstein concentration inequality in Lemma \ref{lem:emp_bern} below. 

\begin{lemma}[Empirical Bernstein Inequality \cite{Mnih2008EmpiricalBS}]\label{lem:emp_bern}
    Let $X_1,\dots,X_n$ be i.i.d random variables with range $R$. With probability at least $1-\delta$, 
    \begin{equation}
        \mu \geq \bar{X} - \bar{\sigma}_n \sqrt{\frac{2\log(\frac{3}{2\delta})}{n}} + \frac{3R\log(\frac{3}{2\delta})}{n}
    \end{equation}
    
\end{lemma}

Note that $\hat{V}_j^{IPW}(\pi) - (1+w_j)\hat{V}_j^{IPW}(\pi_0)$ can be re-expressed as $n$ i.i.d. terms with range $R = |2+w_j|$. By union bounding over the fixed set $\widetilde{\Pi}$ and outcomes $j \in \Scal$, we immediately obtain the results of Lemma \ref{lem:finite_sample_joint_ci} using the concentration inequality in Lemma \ref{lem:emp_bern}. 

\paragraph{Proof of Lemma \ref{lem:asymp_err_control}} The proof of Lemma \ref{lem:asymp_err_control} follows from an asymptotic normality result for doubly robust estimation \cite{dml}. We provide an abridged version of these results with sufficient conditions in Lemma \ref{lem:dml_normal} for our setup in Section \ref{sec:setup}.

\begin{lemma}[Asymptotic Consistency and Inference \cite{dml}]\label{lem:dml_normal}
    Assume that (i) estimated functions $\hat{\mu}_l$, $\hat{e}_l$ are bounded with respect to $P$ almost surely, (ii) $\|\hat{\mu}_l - \mu \|_{P, 2} \times \| \hat{e}_l - e\|_{P, 2} = o_P(1/\sqrt{n})$, and (iii) $\text{Var}(Y|A=a,X=x)$ be bounded below for all $a \in [K], X=x$. Let $\hat{D}_j(\pi)$ and $\hat{d}_j(O, \pi)$ be as defined in Definition \ref{defn:joint_asymp_lower_bds}, and $D_j(\pi), d_j(O, \pi)$ denote these functions with the true nuisances (i.e., true conditional regression functions $\mu$, true propensities $e$). Let $D(\pi)$ denote the $|\Scal|$-length vector $[D_j(\pi)]_{j\in\Scal}$, with $D(\pi)$ defined analogously. 
    
    Then, $\hat{D}(\pi)$ is asymptotically normal, i.e.,
    $$ \sqrt{n}  (\hat{D}(\pi) - D(\pi)) \xrightarrow[d]{} N(0, \Sigma), $$
    where $\Sigma_{ij}$, the $ij$-th entry of $\Sigma$, equals:
    $${\Sigma}_{ij} = \EE_P[\left(\hat{d}_i(O, \pi)-D_i(\pi) \right)\left(\hat{d}_j(O, \pi) - D_j(\pi) \right)],$$. Furthermore, the estimated covariance matrix converges in probability to the true covariance matrix, i.e., $\hat{\Sigma}\xrightarrow{p}\Sigma$.   
\end{lemma}

Using the results of Lemma \ref{lem:dml_normal}, Lemma \ref{lem:asymp_err_control} directly follows by considering the combined $|\Scal| \times |\widetilde{\Pi}|$ vector with entries populated by vectors $\hat{D}(\pi_1), ... ,\hat{D}(\pi_{|\widetilde{\Pi}|})$. By direct application of Equations 5 and 6 for the sup-$t$ band \cite{sup_t_band}, we obtain the desired result for Lemma \ref{lem:asymp_err_control}. 

\paragraph{Proof of Lemma \ref{lem:post_selection_correction}}

This proof follows from the definition of $\epsilon$-stability presented in Definition \ref{defn:stable} and an existing result from the differential privacy literature, provided below.

\begin{lemma}[Max Information Bound \cite{bassily2015algorithmicstabilityadaptivedata}]\label{lem:max_info_bound}
    Define the max information $I^\delta(\widehat{\Pi}; \{O_i\}_{i=1}^n)$ as the following ratio:
    $$I^\delta(\widehat{\Pi}; \{O_i\}_{i=1}^n) = \max_{\Ecal} \log\frac{P\left( (\widehat{\Pi}, \{O_i\}_{i=1}^n) \in \Ecal\right)-\delta}{P\left((\widehat{\Pi},\{O_i'\}_{i=1}^n) \in \Ecal\right)}, $$
    where $\{O'_i\}_{i=1}^n$ is an i.i.d. copy of $\{O_i\}_{i=1}^n$, and $\Ecal$ is any measurable event. Suppose an algorithm $\Acal$ is $\epsilon$-stable, and fix $\delta \in (0,1)$. Then, the max information $I^\delta$ is bounded as follows:
    $$I^\delta(\widehat{\Pi}; \{O_i\}_{i=1}^n) \leq \frac{n}{2}\epsilon^2 + \epsilon\sqrt{\frac{n\log(2/\delta)}{2}}. $$
\end{lemma}

Using Lemma \ref{lem:max_info_bound}, we construct our post-selection adjustments. Define $\Xi$ as the event that any of our finite sample lower bounds in Lemma \ref{lem:finite_sample_joint_ci} are above the true value of weighted policy differences across all $\pi \in \widehat{\Pi}$, where $\widehat{\Pi}$ is obtained through an $\epsilon$-stable algorithm. The $\Xi$ is measurable. Then, by direct application of Lemma \ref{lem:max_info_bound}, 

\begin{align*}
    P((\widehat{G}, \{O_i\}_{i=1}^n \in \Xi) &\leq \exp\left(I^\delta(\widehat{\Pi};\{O_i\}_{i=1}^n)\right) P\left( (\widehat{G}, \{O'_i\}_{i=1}^n) \in \Xi\right) + \delta \\
    &\leq \exp\left(\frac{n}{2}\epsilon^2 + \epsilon\sqrt{\frac{n\log(2/\delta)}{2}} \right)\alpha + \delta
\end{align*}
where the second line holds due to $\{O_i'\}_{i=1}^n$ being an i.i.d. copy of $\{O_i\}_{i=1}^n$. By solving for $\alpha$, which is input error tolerance for the post-selection joint confidence bounds being larger than any true population value in Lemma \ref{lem:finite_sample_joint_ci}, we obtain our expression for $\alpha(\delta')$ in Lemma \ref{lem:post_selection_correction}. We obtain our asymptotic guarantees by applying the reverse Fatou lemma to the argument above, which completes our proof of Lemma \ref{lem:post_selection_correction}.

\paragraph{Proof of Theorem \ref{thm:safe_policy_search}} 

Our proof follows from first noting that our policy pruning approach is a direct application of the Sparse Vector Technique (SVT) \cite[Algorithm 1 of Section 3.6]{dwork_textbook} from the differential privacy literature. We formalize this with Lemma \ref{lem:svt} below.

\begin{lemma}[$\epsilon$-Stability of Sparse Vector Technique \cite{dwork_textbook}]\label{lem:svt}
    Define $\Delta_S$ as the sensitivity of a scoring function $S: \Pi \times \{O_i\}_{i=1}^n \rightarrow \RR$, which is defined as  
    $$\Delta_S = \max_{\pi \in \Pi}\max_{\{O'_i\}_{i=1}^n \simeq \{O_i\}_{i=1}^n }| S\left(\pi, \{O'_i\}_{i=1}^n\right) - S\left(\pi, \{O_i\}_{i=1}^n\right)  |,$$
    where $\{O'_i\}_{i=1}^n \simeq \{O_i\}_{i=1}^n $ is used to denote two sets of $n$ observations that differ by at most one entry.
    
    The Sparse Vector Technique with noise $\text{Lap}(2\eta \Delta_S/\epsilon)$ added to the threshold and noise $\text{Lap}(4\eta\Delta_S/\epsilon)$ added to $S(\pi, \{O_i\}_{i=1}^n)$ is $\epsilon$-stable. 
\end{lemma}

Given this result, it suffices to show that the scoring function used in Algorithm \ref{alg:safe_policy_learning}'s application of SVT. The scoring function in Algorithm \ref{alg:safe_policy_learning} is the minimum of estimated lower bounds. Note that by perturbing one entry in the observed data for any possible dataset realization, any lower bound for a policy $\pi \in \Pi$ across outcomes $j \in \Scal$ can change by at most $B_\text{finite}$ or $B_\text{asymptotic}$ with joint lower bounds constructed with Lemma \ref{lem:finite_sample_joint_ci} or Definition \ref{defn:joint_asymp_lower_bds} respectively. In the asymptotic case, this bound overestimates the change, as $\Phi^{-1}(1-\frac{\alpha'(\delta^*)}{\eta |\Scal|})$ is larger than the sup-$t$ critical value corresponding to $\alpha'(\delta^*)$. Note that if an algorithm is $\epsilon$-stable, then it must also be $\epsilon'$-stable for all $\epsilon' \geq \epsilon$ by Definition \ref{defn:stable}. Thus, using an upper bound on the sensitivity does not affect our results. Therefore, by direct application of Lemma \ref{lem:svt} and noting that our pruning approach takes in an upper bound on the scoring function sensitivity, we obtain the results of Theorem \ref{thm:safe_policy_search}.

\section{Additional Experimental Details}\label{sec:appendix_exp}
We provide additional details regarding (i) further clarifications on Algorithm \ref{alg:safe_policy_learning}, (ii) experimental baselines and (iii) a simple synthetic experiment that demonstrates that SNPL performs best relative to our tested baselines in  low SNR regimes. 

\subsection{Further Details on Algorithm \ref{alg:safe_policy_learning}}

We expand on the potential confusions regarding Algorithm \ref{alg:safe_policy_learning} discussed in footnote \ref{footnote:alg1}. To obtain $(\Scal, \alpha,w)$-asymptotic safety guarantees, Algorithm \ref{alg:safe_policy_learning} uses the lower bounds in \ref{defn:joint_asymp_lower_bds}. These lower bounds depend on the final policy set $\widehat{\Pi}$, which is unknown before the loop in Algorithm \ref{alg:safe_policy_learning} terminates. As a heuristic, we use the current set $\widehat{\Pi}$ within the loop to construct the lower confidence bounds. Note that this results in optimistic lower bound estimates within the loop, necessitating lower bound re-estimation in line 14. This re-estimation is still necessary when we use the finite-sample variant. For the finite sample variant, we assume $|\widehat{\Pi}| = \eta$ within the loop when constructing our joint lower bounds according to Lemma \ref{lem:finite_sample_joint_ci} for the policy pruning loop. This leads to potentially conservative lower bound estimates. As such, line 14 is still necessary.

\subsection{Additional Details for Experiment Baselines}

Here, we provide pseudocode regarding the baselines, and justify their safety guarantees. We begin with our data-splitting approach. 
\begin{algorithm}[t!] 
  \caption{Data Splitting Approach (HCPI \cite{pmlr-v37-thomas15})} \label{alg:ds}
  \begin{algorithmic}[1] 
  \State 
  \textbf{input}: data 
  $\{O_i\}_{i=1}^n$, goal outcome $g$, guardrail outcomes $\Scal$, weight vector $w$, error level $\alpha$, policy class $\Pi$, split fraction $\rho$, baseline policy $\pi_0$,lower confidence bound construction method $\hat{C}$, goal policy value estimator $\hat{V}_g$. 

  \State Split the data $\{O_i\}_{i=1}^n$, with $\rho$ fraction of the data in learning split $L$ and $1-\rho$ fraction of the data in testing split $T$. 

  \For{$\pi \in \Pi \setminus \{\pi_0\}$}

  \State For $j \in \Scal$, estimate joint bounds $\hat{C}\left(\pi, \alpha'(\delta^*)\right)$ with $L$.

  \State Set $M'(\pi) = \min_{j \in \Scal} \hat{C}_j\left(\pi, \alpha'(\delta^*)\right)$.

  \State Estimate $\hat{V}_g^L(\pi)$ with learning split $L$. 
  \EndFor

  \State Select $\hat{\pi} = \argmax_{\pi \in \Pi} f(\pi)$, where
  $$f(\pi) = \mathbf{1}[M'(\pi)\geq 0] \ \hat{V}_g^L(\pi) + \mathbf{1}[M'(\pi) < 0] M'(\pi).$$

  \State Estimate $\hat{M}(\hat\pi) = \min_{j \in \Scal} \hat{C}_j(\pi, \alpha)$ with test split $T$.

  \If{$\hat{M}(\hat\pi) > 0 $} return $\hat\pi$. 
  \Else{ \ return  $\pi_0$.}
  \EndIf

\label{alg:data_splitting}
  \end{algorithmic}    
\end{algorithm}

\paragraph{Data-Splitting Approach} Our data-splitting directly adapts the High Confidence Policy Improvement (HCPI) algorithm \cite{pmlr-v37-thomas15} intended for a single outcome, and generalizes it to the multi-outcome setting. We provide the pseudocode for this approach in Algorithm \ref{alg:ds}. First, we split the data into two partitions based on a pre-specified fraction $\rho$, with a pre-specified $\rho$-fraction of the data allocated for learning, and its complement allocated for testing. We estimate joint lower confidence bounds and goal policy values on the learning set, and select $\hat\pi$ as the policy with the largest estimated value in goal metric $g$ among those whose estimated joint lower bounds are above zero. If no policies are estimated to pass, then we select $\hat\pi$ as the policy with the largest estimated minimum lower bound on the learning dataset. Given the policy $\hat\pi$, we then test that its minimum lower bound is above zero on the test data. If this holds, then we return the policy $\hat\pi$; if not, then we revert to the baseline policy $\pi_0$. This approach maintains safety by (i) using an distinct sets of data for learning and testing and (ii) reverting to the baseline $\pi_0$ if \emph{any} lower bounds for the guardrail policy values fall below zero. We obtain an $(\Scal, w, \alpha)$-safe algorithm using the estimator $\hat{V}_g$ in Definition \ref{defn:ipw} and lower bounds in Lemma \ref{lem:finite_sample_joint_ci}, and an $(\Scal, w, \alpha)$-asymptotically safe algorithm using the estimator $\hat{V}_g$ in Definition \ref{defn:dml} and lower bounds in Definition \ref{defn:joint_asymp_lower_bds}.

\paragraph{Bonferroni Approach} Our Bonferroni approach simply uses the Bonferroni correction \cite{Dunn_1961} on the joint lower bounds across all policies $\pi \in \Pi$. We then select the policy with the largest policy value $\hat{V}_g$ among those that have all corrected joint lower bounds (i.e., minimum corrected joint lower bounds) above zero. Like our data-splitting approach, we obtain an $(\Scal, w, \alpha)$-safe algorithm using the estimator $\hat{V}_g$ in Definition \ref{defn:ipw} and lower bounds in Lemma \ref{lem:finite_sample_joint_ci}, and an $(\Scal, w, \alpha)$-asymptotically safe algorithm using the estimator $\hat{V}_g$ in Definition \ref{defn:dml} and lower bounds in Definition \ref{defn:joint_asymp_lower_bds}. The safety guarantees directly follow from a simple union bound argument and the results of Lemma \ref{lem:finite_sample_joint_ci} (for finite sample safety guarantees) or Lemma \ref{lem:asymp_err_control} (for asymptotic safety guarantees).

\begin{table}[t!]
    \centering    \caption{Detection Rate and EI on synthetic data.}
    \label{tab:sample_table}
    \begin{tabular}{c | c|c|c}
      &  Method & Detection Rate & EI \\ \hline
   &    DS (25\%) & 0.513 $\pm$ 0.056 & \textbf{0.051 $\pm$ 0.006} \\ 
      &  DS (50\%) & 0.396 $\pm$ 0.055 & 0.040 $\pm$ 0.006 \\ 
    $|\Pi| = 2500$  &  DS (75\%) & 0.206 $\pm$ 0.046 & 0.018 $\pm$ 0.004 \\ 
      &  Bonferroni & \textbf{0.543 $\pm$ 0.056} & 0.036 $\pm$ 0.005 \\ 
      &  SNPL & 0.200 $\pm$ 0.045 & 0.024$\pm$ 0.005 \\ \hline
     &   DS (25\%) & 0.513 $\pm$ 0.056 & \textbf{0.049 $\pm$ 0.006} \\ 
      &  DS (50\%) & 0.380 $\pm$ 0.054 & 0.037 $\pm$ 0.006 \\ 
      $|\Pi|= 1000$ & DS (75\%) & 0.223 $\pm$ 0.047 & 0.021 $\pm$ 0.004 \\ 
     &   Bonferroni & \textbf{0.540 $\pm$ 0.056} & 0.043 $\pm$ 0.006 \\ 
     &   SNPL & 0.226 $\pm$ 0.047 & 0.026 $\pm$ 0.005 \\ 
      \hline
    & DS (25\%) & 0.466 $\pm$ 0.056 & 0.046 $\pm$ 0.006 \\ 
       & DS (50\%) & 0.400 $\pm$ 0.055 & 0.038$\pm$ 0.006 \\ 
    $|\Pi| = 500$   & DS (75\%) & 0.206$\pm$ 0.046 & 0.017 $\pm$ 0.004 \\ 
       & Bonferroni & \textbf{0.626 $\pm$ 0.055} & \textbf{0.050 $\pm$ 0.005} \\ 
       & SNPL & 0.186 $\pm$ 0.044 & 0.021 $\pm$ 0.005 \\ 
    \end{tabular}
\end{table}

\subsection{Synthetic Experiments for High SNR}

\begin{figure*}[h]
    \centering
    \includegraphics[width=0.3\linewidth]{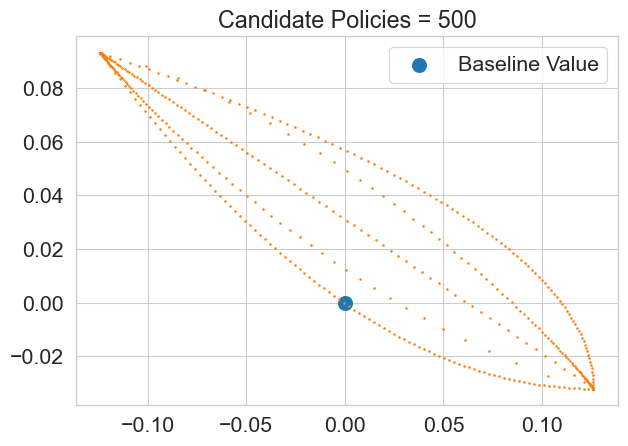}\Description{}
    \includegraphics[width=0.3\linewidth]{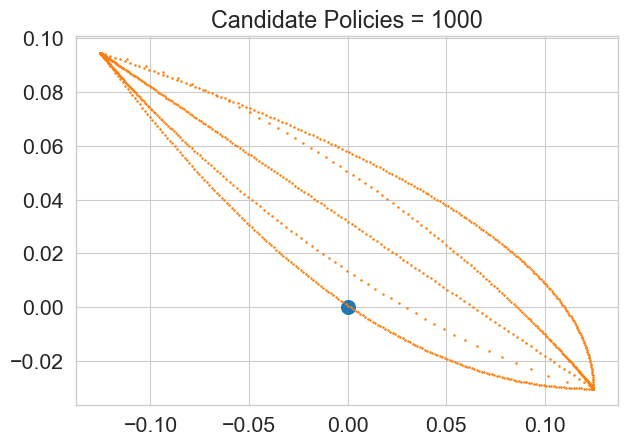}\Description{}
    \includegraphics[width=0.3\linewidth]{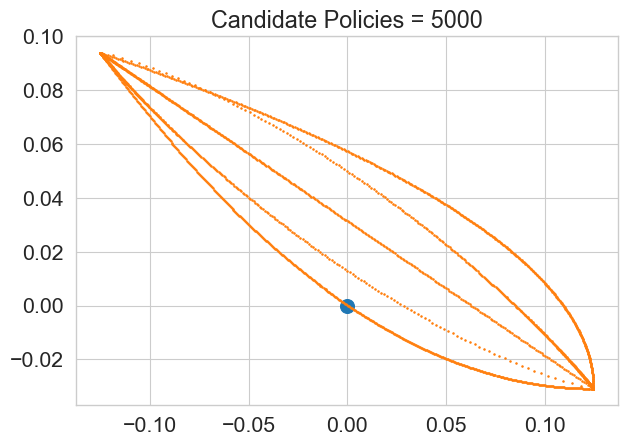}\Description{}
    \caption{Plots of Policy Values over Candidate Policy Class. $X$-axis represents policy values for outcome 1, while $Y$-axis represents policy values for outcome 2. All values are centered using the respective baseline policy values for each outcome.}
    \label{fig:policy_class_plot}
\end{figure*}

We design a straightforward synthetic example to emphasize that our method is deliberately designed for \emph{low} SNR regimes with \emph{large} policy classes. We set $\gamma = 0.1$, $\alpha=0.1$, and $p=0.5$, which gives us the choices for $\epsilon$ and $\eta$ in Algorithm \ref{alg:safe_policy_learning} according to Table \ref{tab:hypereparameters}. Our data-generating process consists of covariates $X_1,X_2,X_3 \sim \text{Unif}(0,1)$, randomized treatments $A \sim \text{Bern}(0.5)$, and two binary outcomes $Y_1 \sim \text{Bern}(f_1(A, X))$, $Y_2 \sim \text{Bern}(f_1(A,X))$,
$$f_1(A,X) = 0.5(1-\mathbf{1}[A=1]X_2), \ f_2(A,X) = 0.5(1+\mathbf{1}[A=1] X_1X_3 ).$$ 
We set our goal outcome as \textit{outcome 1}. Our policy class consists of threshold policies based on covariates $X$, taking the form
$$\pi(X) = \mathbf{1}[g_i(X) < c],$$
where we use the following choices of $g_i$: $g_1(X) = X_1$, $g_2(X)  = X_2$, $g_3(X) = X_1X_2$, $g_4(X) = X_1X_2X_3$, and $g_5(X) = -X_1X_2X_3$. For each $g_i$, we vary values of the cutoff $c$ over a uniformly spaced grid of values between $[0,1]$. We test policy classes with the size $|\Pi| \in \{500, 1000, 2500\}$ by setting the grid of cutoffs $c$ as $100$, $200$, and $500$ evenly spaced points over this range. As our baseline across all experiments, we set $\pi_0(X) = \mathbf{1}[X_1 < 0.5]$. To visualize the policy values for each candidate policy set, we plot the distribution of policy values, normalized by the baseline policy value, in Figure \ref{fig:policy_class_plot}. For this simple synthetic experiment we use approaches with asymptotic safety guarantees. For our outcome regressions $\hat\mu$ in Definition \ref{defn:dml}, we use linear regression with $k=5$ folds as described in Definition \ref{defn:dml}. To demonstrate our current heuristics for SNPL are primarily intended for low SNR regimes, we test higher SNR ratios in our synthetic example with the same $n=1000$, where we see the most benefits on real-world data with asymptotic approaches. This set-up provides high SNR ratios relative to the experimental data in the case study. The minimum SNR ratio is 0.137 (compared to roughly $0.002$ for our real-world example at the same sample size). We set $w = [0, -0.1]$, indicating that acceptable policies must achieve at least 90\% of the baseline value for outcome 2 while strictly improving on outcome 1. For all reported values, we obtain these estimates over 300 simulated runs for each approach and policy class combination.

\paragraph{Discussion of Synthetic Results} 

 We report the detection rates and EI in Table \ref{tab:sample_table}. For all methods, Type I error is zero. Our results with $|\Pi| = 2500$ and $|\Pi| = 1000$ demonstrate the effect of significantly higher SNR in large and moderately sized policy classes. In these settings, DS (25\%) reports the largest expected improvement, while Bonferroni provides the largest detection rates. SNPL only achieves roughly 40\% of the best-performing option for both detection rate and EI in this high SNR regime. Our results with $|\Pi| = 500$ demonstrate that the relative performance of SNPL degrades with a small number of candidate policies. With $|\Pi| = 500$, SNPL's detection rates drop to roughly 30\% to that on Bonferroni, the best performing option. With large policy classes and high SNR, data-splitting with a small proportion of data (i.e., 25\%) used to select the policy performs favorably. With small policy classes and high SNR, Bonferroni approaches lead to the largest levels of detection rate and EI. Despite not being designed for this setting, SNPL still achieves similar, if not better, performance than some baselines. For each policy set size $|\Pi|$, SNPL achieves similar detection rates to DS (75\%), while achieving strictly larger values of EI. To improve performance, we plan to investigate different heuristics for SNPL's hyperparameters across different SNR regimes in future work.

\end{document}